# Circular Pythagorean fuzzy sets and applications to multi-criteria decision making

Mahmut Can Bozyiğit [1], Murat Olgun [2], Mehmet Ünver [2,*]

[1]Ankara Yıldırım Beyazıt University, Faculty of Engineering and Natural Sciences,
Department of Mathematics, 06420 Ankara Türkiye
mcbozyigit@ybu.edu.tr
[2]Ankara University, Faculty of Science, Department of Mathematics,
06100 Ankara Türkiye
olgun@ankara.edu.tr, munver@ankara.edu.tr

## Abstract

In this paper, we introduce the concept of circular Pythagorean fuzzy set (value) (C-PFS(V)) as a new generalization of both circular intuitionistic fuzzy sets (C-IFSs) proposed by Atannassov and Pythagorean fuzzy sets (PFSs) proposed by Yager. A circular Pythagorean fuzzy set is represented by a circle that represents the membership degree and the non-membership degree and whose center consists of non-negative real numbers $\mu$ and $\nu$ with the condition $\mu^2 + \nu^2 \leq 1$. A C-PFS models the fuzziness of the uncertain information more properly thanks to its structure that allows modelling the information with points of a circle of a certain center and a radius. Therefore, a C-PFS lets decision makers to evaluate objects in a larger and more flexible region and thus more sensitive decisions can be made. After defining the concept of C-PFS we define some fundamental set operations between C-PFSs and propose some algebraic operations between C-PFVs via general $t$-norms and $t$-conorms. By utilizing these algebraic operations, we introduce some weighted aggregation operators to transform input values represented by C-PFVs to a single output value. Then to determine the degree of similarity between C-PFVs we define a cosine similarity measure based on radius. Furthermore, we develop a method to transform a collection of Pythagorean fuzzy values to a PFS. Finally, a method is given to solve multi-criteria decision making problems in circular Pythagorean fuzzy environment and the proposed method is practiced to a problem about selecting the best photovoltaic cell from the literature. We also study the comparison analysis and time complexity of the proposed method.

**Keywords:** Circular Pythagorean fuzzy set, aggregation operators, multi-criteria decision making

## 1   Introduction

The concept of fuzzy set (FS) was developed by utilizing a function (called membership function) assigning a value between zero and one as the membership degrees of the elements to deal with ambiguity in real-life problems. Since the FS theory proposed by Zadeh [37] succeeded to handle various types of uncertainty, it has been studied in detail by many researchers to model uncertainty. Later the concept of intuitionistic fuzzy set (IFS), which is an extension of the concept of FS, was proposed by Atanassov [1] via membership functions and non-membership functions. The theory of IFS plays an important role in many research areas such as pattern recognition, multi-criteria decision making (MCDM), data mining, classification, clustering and medical diagnosis. Many aggregation operators, similarity measures, distance measures and entropy measures have been developed for IFSs. Particularly, various generalizations of aggregation operators for IFSs (see e.g. [6, 11]) have been defined via particular types of $t$-norms and $t$-conorms.

The concept of Pythagorean fuzzy set (PFS) that is a ricing tool in MCDM (see, Figure 1) was introduced by Yager [31, 32] to research in a wider environment to express uncertainty as a generalization of the concept of IFS. A PFS is characterized via a membership function and a non-membership function such that the sum of the squares of these non-negative functions are less than 1. Moreover, a PFS has a quadratic form, which means a PFS expands the range of the change of membership degree and non-membership degree to the unit circle and so is more capable than an IFS in depicting uncertainty. Yager [32, 33] proposed some aggregation operators for PFSs. After that, Peng et. al. [25] presented the axiomatic definitions of distance measure, similarity measure and entropy measure for PFSs. Further studies on MCDM with fuzzy sets and aggregation operators can be found in [5, 7, 10, 11, 12, 22, 23, 24, 28, 29, 35, 36].

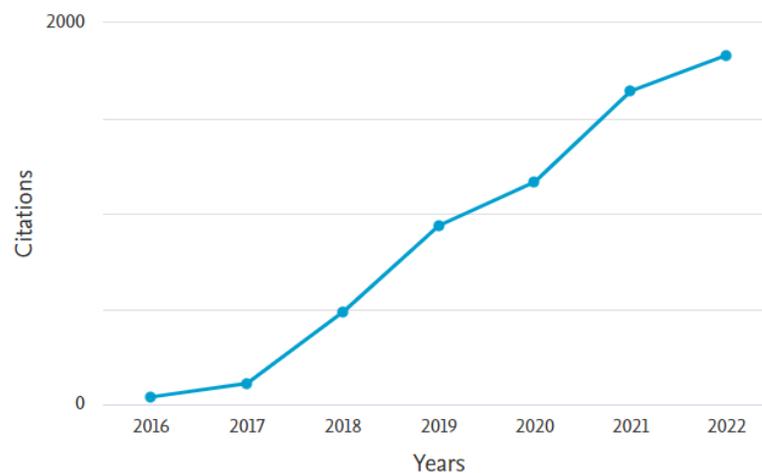

Figure   1: Citation graph of the PFSs

Many types of fuzzy sets study with points, pairs of points or triples of points from the closed interval [0,1] that makes the decision process more strict since they require (decision makers) DMs to assign precise numbers. To overcome such a strict modelling Atanassov [2] proposed the concept of circular intuitionistic fuzzy set (C-IFSs). A C-IFS is represented by a circle standing for the uncertainty of the membership and non-membership functions. That is, the membership and the non-membership of each element to a C-IFS are shown as a circle whose center is a pair of non-negative real numbers with the condition that the sum of them is less than 1. With the help of C-IFSs, the change of membership degree and non-membership degree can be handled more sensitively to express uncertainty. Therefore, various types of MCDM methods have been carried to circular intuitionistic fuzzy environment (see e.g. [3, 15, 16]). In this paper, we carry the idea of representing membership degree and non-membership degree as circle to the Pythagorean fuzzy environment by introducing the concept of circular Pythagorean fuzzy set (C-PFS). In this new fuzzy set notion, the membership and non-membership degrees of an element to a FS are represented by circles with center $(\mu_A(x), v_A(x))$ instead of numbers and with a more flexible condition $\mu_A^2(x) + v_A^2(x) \leq 1$. In this manner, we extend not only the concept of the PFS, but also the concept of the C-IFS (see Figure 1). Thus the decision making process become more sensitive since DMs can attain circles with certain properties instead of precise numbers. Figure 1 illustrates the improvement of circular fuzzy sets.

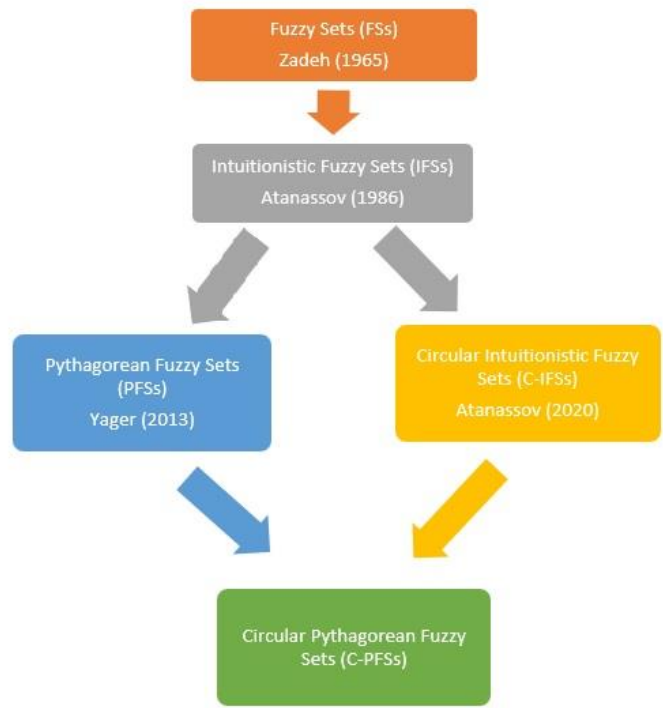

Figure 2: The improvement of circular fuzzy set theory

Some main contributions of the present paper can be given as follows.

• This paper introduces the concepts of C-PFS and circular Pythagorean fuzzy value (C-PFV).

• A method is developed to transform a collection of Pythagorean fuzzy values (PFVs) to a C-PFS. In this way, multi-criteria group decision making (MCGDM) problem can be relieved.

• The membership and non-membership of an element to a C-PFS are represented by circles. Thanks to its structure a more sensitive modelling can be done in MCDM theory in the continuous environment.

• Some algebraic operations are defined for C-PFVs via $t$-norms and $t$-conorms. With the help of these operations some weighted arithmetic and geometric aggregation operators are provided. These aggregation operators are used in MCDM and MCGDM.

The rest of this paper is organized as follows. In Section 2, we recall some basic concepts. In Section 3, we introduce the concept of C-PFS(V) as new generalization of both C-IFSs and PFSs. We also define some fundamental set theoretic operations for C-PFSs. Then we introduce some algebraic operations for C-PFVs via continuous Archimedean $t$- norms and $t$- conorms. In Section 4, we propose some weighted aggregation operators for C-PFVs by utilizing these algebraic operations. In Section 5, motivating by a cosine similarity measure defined for PFVs in [30], we define a cosine similarity measure for C-PFVs to determine the degree of similarity between C-PFVs. Using the proposed similarity measure and the aggregation operators we provide a MCDM method in circular Pythagorean fuzzy environment. We also apply the proposed method to a MCDM problem from the literature [39] that deals with selecting the best photovoltaic cell (also known as solar cell). We compare the results of the proposed method with the existing result and calculate the time complexity of the MCDM method. In Section 6, we conclude the paper.

## 2 Preliminaries

Atanassov [1] introduced the concept of IFS by taking into account the non-membership functions with a membership functions of FSs. Throughout this section we assume that $X = \{x_1, \ldots, x_n\}$ is a finite set.

**Definition 1** *[1] An IFS $A$ in $X$ is defined by*
$$A = \{\langle x, \mu_A(x), \nu_A(x)\rangle : x \in X\}$$
where $\mu_A, \nu_A : X \to [0,1]$ are functions with the condition
$$\mu_A + \nu_A \leq 1$$
that are called the membership function and the non-membership function, respectively.

The concept of PFS proposed by Yager [31, 32] which is a generalization of IFS.
**Definition 2** *[31, 32] A PFS $A$ in $X$ is defined by*
$$A = \{\langle x, \mu_A(x), \nu_A(x)\rangle : x \in X\}$$
where $\mu_A, \nu_A : X \to [0,1]$ are functions with the condition
$$\mu_A^2(x) + \nu_A^2(x) \leq 1$$
that are called the membership function and the non-membership function, respectively. Let $\mu_\alpha, \nu_\alpha \in [0,1]$ such that $\mu_\alpha^2 + \nu_\alpha^2 \leq 1$. Then the pair $\alpha = \langle \mu_\alpha, \nu_\alpha \rangle$ is called a Pythagorean fuzzy value (PFV).

Schweizer and Sklar [26] introduced the concepts of $t$-norm and $t$-conorm by motivating the concept of probabilistic metric spaces proposed by Menger [21]. These concepts have important roles in statistic and decision making. Algebraically, $t$-norms and $t$-conorms are binary operations defined on the closed unit interval.

**Definition 3** *[17, 26] A $t$-norm is a function $T : [0,1] \times [0,1] \to [0,1]$ that satisfies the following conditions:*

(T1) $T(x, 1) = x$ for all $x \in [0,1]$ (border condition),
(T2) $T(x, y) = T(y, x)$ for all $x, y \in [0,1]$ (commutativity),
(T3) $T(x, T(y, z)) = T(T(x, y), z)$ for all $x, y, z \in [0,1]$ (associativity),
(T4) $T(x, y) \leq T(x', y')$ whenever $x \leq x'$ and $y \leq y'$ for all $x, x', y, y' \in [0,1]$ (monotonicity).

**Definition 4** *[17, 26] A t-conorm is a function $S: [0,1] \times [0,1] \to [0,1]$ that satisfies the following conditions:*

(S1) $S(x, 0) = x$ for all $x \in [0,1]$ (border condition),
(S2) $S(x, y) = S(y, x)$ for all $x, y \in [0,1]$ (commutativity),
(S3) $S(x, S(y, z)) = S(S(x, y), z)$ for all $x, y, z \in [0,1]$ (associativity),
(S4) $S(x, y) \leq S(x', y')$ whenever $x \leq x'$ and $y \leq y'$ for all $x, x', y, y' \in [0,1]$ (monotonicity).

**Definition 5** *[17, 18] A strictly decreasing function $g: [0,1] \to [0, \infty]$ with $g(1) = 0$ is called the additive generator of a t-norm $T$ if we have $T(x, y) = g^{-1}(g(x) + g(y))$ for all $(x, y) \in [0,1] \times [0,1]$.*

Next, we need the concept of fuzzy complement to find the additive generator of a dual t-conorm on $[0,1]$.

**Definition 6** *[31, 32, 34] A fuzzy complement is a function $N: [0,1] \to [0,1]$ satisfying the following conditions:*

(N1) $N(0) = 1$ and $N(1) = 0$ (boundary conditions),
(N2) $N(x) \geq N(y)$ whenever $x \leq y$ for all $x, y \in [0,1]$ (monotonicity),
(N3) Continuity,
(N4) $N(N(x)) = x$ for all $x \in [0,1]$ (involution).

The function $N: [0,1] \to [0,1]$ defined by $N(a) = (1 - a^p)^{1/p}$ where $p \in (0, \infty)$ [31, 32] is a fuzzy complement. When $p = 2$, $N$ becomes the Pythagorean fuzzy complement $N(a) = \sqrt{1 - a^2}$.

**Definition 7** *[20, 34] Let $T$ be a t-norm and let $S$ be a t-conorm on $[0,1]$. If $T(x, y) = N(S(N(x), N(y)))$ and $S(x, y) = N(T(N(x), N(y)))$, then $T$ and $S$ are called dual with respect to a fuzzy complement $N$.*

**Remark 1** *Let $T$ be a t-norm on $[0,1]$. Then the dual t-conorm $S$ with respect to the Pythagorean fuzzy complement $N$ is*

$$S(x, y) = \sqrt{1 - T^2(\sqrt{1 - x^2}, \sqrt{1 - y^2})}.$$

Note that $T$ is an Archimedean $t$-norm if and only if $T(x,x) < x$ for all $x \in (0,1)$ and $S$ is an Archimedean $t$-conorm if and only if $S(x,x) > x$ for all $x \in (0,1)$ [17, 18]. Klement and Mesiar [19] proved that continuous Archimedean $t$-norms have representations via their additive generators in the following theorem.

**Theorem 1** *[19] Let $T$ be a $t$-norm on $[0,1]$. The following statements are equivalent:*

(i) $T$ is a continuous Archimedean $t$-norm.

(ii) $T$ has a continuous additive generator, i.e., there is a continuous, strictly decreasing function $g: [0,1] \to [0,\infty]$ with $t(1) = 0$ such that $T(x,y) = g^{-1}(g(x) + g(y))$ for all $(x,y) \in [0,1] \times [0,1]$.

## 3 Circular Pythagorean fuzzy sets

The notion of C-IFS was introduced by Atanassov [2] as an extension of the notion of IFS. Throughout this paper we assume that $X = \{x_1, \ldots, x_n\}$ is a finite set.

**Definition 8** *[2] Let $r \in [0,1]$. A circular C-IFS $A_r$ in X is defined by*
$$A_r = \{\langle x, \mu_A(x), \nu_A(x); r\rangle : x \in X\}$$
where $\mu_A, \nu_A : X \to [0,1]$ are functions such that
$$\mu_A + \nu_A \leq 1.$$
$r$ is the radius of the circle around the point $(\mu_A(x), \nu_A(x))$ on the plane. This circle represents the membership degree and non-membership degree of $x \in X$.

**Remark 2** *Since each IFS $A$ has the form*
$$A = A_0 = \{\langle x, \mu_A(x), \nu_A(x); 0\rangle : x \in X\}$$
any IFS can be considered as a C-IFS. Hence, the notion of C-IFS is a generalization of the notion of IFS.

Next we introduce the concept of C-PFS that is a new extension of the concepts of C-IFS and PFS. C-PFSs allow decision makers to express uncertainty via membership and non-membership degrees represented by a circle in a more extended environment. Thus more sensitive evaluations can be made in decision making process.

**Definition 9** *Let $r \in [0,1]$. A C-PFS $A_r$ in X is defined by*
$$A_r = \{\langle x, \mu_A(x), \nu_A(x); r\rangle : x \in X\}$$
where $\mu_A, \nu_A : X \to [0,1]$ are functions such that
$$\mu_A^2 + \nu_A^2 \leq 1.$$
$r$ is the radius of the circle around the point $(\mu_A(x), \nu_A(x))$ on the plane. This circle represents the membership degree and non-membership degree of $x \in X$.

**Example 1** *Let $X = \{x_1, x_2, x_3\}$. An example of a C-PFS on $X$ can be given by*

$$A_{0.2} = \{\langle x_1, 0.3, 0.8; 0.2\rangle, \langle x_2, 0.1, 0.9; 0.2\rangle, \langle x_3, 0.5, 0.6; 0.2\rangle\}.$$

**Definition 10** *Let $\mu_\alpha, \nu_\alpha \in [0,1]$ such that $\mu_\alpha^2 + \nu_\alpha^2 \leq 1$ and $r_\alpha \in [0,1]$. Then the triple $\alpha = \langle \mu_\alpha, \nu_\alpha; r_\alpha \rangle$ is called a C-PFV.*

A C-PFS can be considered as a collection of C-PFVs. Figure 3 shows some examples of C-PFVs and Figure 4 shows that the concept of C-PFS generalizes the concept of C-IFS.

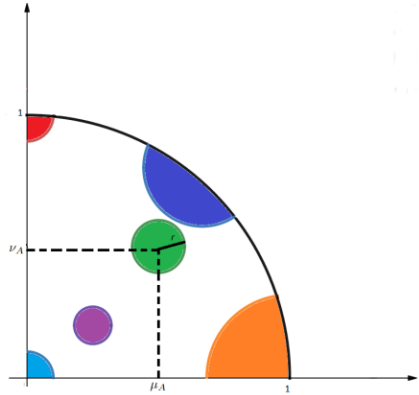

Figure   3: Geometric representation of C-PFSs.

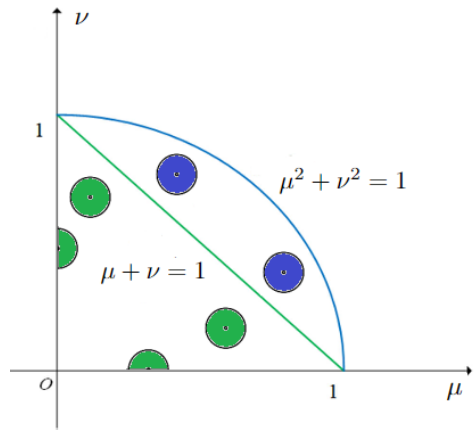

Figure   4: Comparison of the spaces of C-IFS and C-PFS

**Remark 3** *Since each PFS $A$ has the form*
$$A = A_0 = \{\langle x, \mu_A(x), \nu_A(x); 0\rangle : x \in X\},$$
any PFS is also a C-PFS but the converse is not true in general. Consider the C-PFS $A_{0.2}$ given in Example 1. Since $0.3 + 0.8 = 1.1 > 1$ it is not a C-IFS.

Now we can define some set operations among C-PFSs.

**Definition 11** *Let*
$$A_r = \{\langle x, \mu_A(x), \nu_A(x); r\rangle : x \in X\}$$
and
$$B_s = \{\langle x, \mu_B(x), \nu_B(x); s\rangle : x \in X\}$$
be two C-PFSs in $X$. Some set operations among C-PFSs is defined as follows:

a) $A_r \subset B_s$ if and only if $r \leq s$ and

$$\mu_A(x) \leq \mu_B(x) \quad and \quad \nu_A(x) \geq \nu_B(x) \quad for \quad any \quad x \in X.$$

b) $A_r = B_s$ if and only if $r = s$ and

$$\mu_A(x) = \mu_B(x) \quad and \quad \nu_A(x) = \nu_B(x) \quad for \quad any \quad x \in X.$$

c) The complement $A_r^c$ of $A_r$ is defined by $A_r^c = \{\langle x, \nu_A(x), \mu_A(x); r\rangle : x \in X\}$.

d) The union of $A_r$ and $B_r$ with respect to maximum and minimum is defined by
$$A_r \cup_{\min} B_s = \{\langle x, \max(\mu_A(x), \mu_B(x)), \min(\nu_A(x), \nu_B(x)); \min(r, s)\rangle : x \in X\}$$
and
$$A_r \cup_{\max} B_s = \{\langle x, \max(\mu_A(x), \mu_B(x)), \min(\nu_A(x), \nu_B(x)); \max(r, s)\rangle : x \in X\},$$
respectively.

e) The intersection of $A_r$ and $B_r$ with respect to maximum and minimum is defined by
$$A_r \cap_{\min} B_s = \{\langle x, \min(\mu_A(x), \mu_B(x)), \max(\nu_A(x), \nu_B(x)); \min(r, s)\rangle : x \in X\}$$
and
$$A_r \cap_{\max} B_s = \{\langle x, \min(\mu_A(x), \mu_B(x)), \max(\nu_A(x), \nu_B(x)); \max(r, s)\rangle : x \in X\},$$
respectively.

**Example 2** Let $X = \{x_1, x_2, x_3\}$. Consider the C-PFSs given with
$$A_{0.2} = \{\langle x_1, 0.3, 0.8; 0.2\rangle, \langle x_2, 0.1, 0.9; 0.2\rangle, \langle x_3, 0.5, 0.6; 0.2\rangle\}$$
and
$$B_{0.6} = \{\langle x_1, 0.7, 0.5; 0.6\rangle, \langle x_2, 0.2, 0.5; 0.6\rangle, \langle x_3, 0.6, 0.3; 0.6\rangle\}.$$
Then $A_{0.2} \subset B_{0.6}$. On the other hand,
$$A_{0.2}^c = \{\langle x_1, 0.8, 0.3; 0.2\rangle, \langle x_2, 0.9, 0.1; 0.2\rangle, \langle x_3, 0.6, 0.5; 0.2\rangle\},$$

$$A_{0.2} \cup_{\min} B_{0.6} = \{\langle x_1, 0.7, 0.5; 0.2\rangle, \langle x_2, 0.2, 0.5; 0.2\rangle, \langle x_3, 0.6, 0.3; 0.2\rangle\},$$

$$A_{0.2} \cup_{\max} B_{0.6} = \{\langle x_1, 0.7, 0.5; 0.6\rangle, \langle x_2, 0.2, 0.5; 0.6\rangle, \langle x_3, 0.6, 0.3; 0.6\rangle\},$$

$$A_{0.2} \cap_{\min} B_{0.6} = \{\langle x_1, 0.3, 0.8; 0.2\rangle, \langle x_2, 0.1, 0.9; 0.2\rangle, \langle x_3, 0.5, 0.6; 0.2\rangle\},$$
and
$$A_{0.2} \cap_{\max} B_{0.6} = \{\langle x_1, 0.3, 0.8; 0.6\rangle, \langle x_2, 0.1, 0.9; 0.6\rangle, \langle x_3, 0.5, 0.6; 0.6\rangle\}.$$

The following theorem shows that De Morgan's rules are available for C-PFSs.

**Theorem 2** *Let*
$$A_r = \{\langle x, \mu_A(x), \nu_A(x); r\rangle : x \in X\}$$
and

$$B_s = \{\langle x, \mu_B(x), \nu_B(x); s\rangle : x \in X\}$$

be two C-PFSs in $X$. Then we have

1) $(A_r \cup_{\min} B_s)^c = A_r{}^c \cap_{\min} B_s{}^c$
2) $(A_r \cup_{\max} B_s)^c = A_r{}^c \cap_{\max} B_s{}^c$
3) $(A_r \cap_{\min} B_s)^c = A_r{}^c \cup_{\min} B_s{}^c$
4) $(A_r \cap_{\max} B_s)^c = A_r{}^c \cup_{\max} B_s{}^c$

*Proof.* The proof is trivial from Definition 11.

Now we develop a method to convert collections of PFVs to a C-PFS which is a useful method in group decision making.

**Proposition 1** *Let a collection of PFVs is assigned for any $x_i$ by*
$$\{\langle \mu_{i,1}, \nu_{i,1}\rangle, \langle \mu_{i,2}, \nu_{i,2}\rangle, \dots, \langle \mu_{i,k_i}, \nu_{i,k_i}\rangle\}.$$

Then
$$A_r = \{\langle x_i, \mu(x_i), \nu(x_i); r\rangle : x_i \in X\}$$

is a C-PFS where

$$\langle \mu(x_i), \nu(x_i)\rangle = \langle \sqrt{\frac{\sum_{j=1}^{k_i} \mu_{i,j}^2}{k_i}}, \sqrt{\frac{\sum_{j=1}^{k_i} \nu_{i,j}^2}{k_i}}\rangle$$

and
$$r_i = \min\{\max_{1 \leq j \leq k_i} \sqrt{(\mu(x_i) - \mu_{i,j})^2 + (\nu(x_i) - \nu_{i,j})^2}, 1\}.$$

*Proof.* We have

$$\mu^2(x_i) + \nu^2(x_i) = \left(\sqrt{\frac{\sum_{j=1}^{k_i} \mu_{i,j}^2}{k_i}}\right)^2 + \left(\sqrt{\frac{\sum_{j=1}^{k_i} \nu_{i,j}^2}{k_i}}\right)^2$$

$$= \frac{\sum_{j=1}^{k_i} \mu_{i,j}^2}{k_i} + \frac{\sum_{j=1}^{k_i} \nu_{i,j}^2}{k_i}$$

$$= \frac{\sum_{j=1}^{k_i} \mu_{i,j}^2 + \nu_{i,j}^2}{k_i}$$

$$\leq \frac{\sum_{j=1}^{k_i} 1}{k_i}$$

$$= 1.$$

On the other hand it is clear that $0 \leq r_i \leq 1$ for each $i$. Therefore, $A_r = \{\langle x_i, \mu(x_i), \nu(x_i); r\rangle : x_i \in X\}$ is a C-PFS.

**Example 3** Let $X = \{x_1, x_2, x_3\}$ and let collections of PFVs is assigned for any $x_i$ $(i = 1,2,3)$ by

$$\{\langle 0.3, 0.8\rangle, \langle 0.4, 0.6\rangle, \langle 0.5, 0.7\rangle, \langle 0.4, 0.8\rangle\},$$

and
$$\{\langle 0.2, 0.3\rangle, \langle 0.1, 0.4\rangle, \langle 0.2, 0.5\rangle, \langle 0.1, 0.6\rangle\},$$

$$\{\langle 0.9, 0.2\rangle, \langle 0.8, 0.3\rangle, \langle 0.8, 0.2\rangle, \langle 0.7, 0.5\rangle\},$$

respectively. By using Proposition 1, we obtain the C-PFS
$$A = \{\langle x_1, 0.41, 0.73; 0.13\rangle, \langle x_2, 0.16, 0.46; 0.17\rangle, \langle x_3, 0.8, 0.32; 0.2\rangle\}.$$
We visualize this transformation in Figure 5.

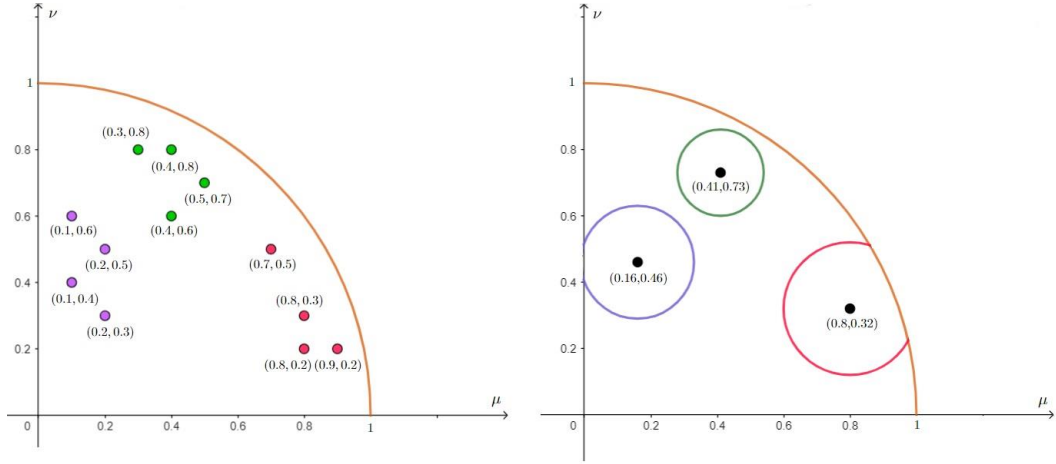

Figure 5: The PFVs and C-PFS in Example 3

Now we define some algebraic operations for C-PFVs.

**Definition 12** Let $\alpha = \langle \mu_\alpha, \nu_\alpha; r_\alpha\rangle$ and $\beta = \langle \mu_\beta, \nu_\beta; r_\beta\rangle$ be two C-PFVs. Some algebraic operations among C-PFVs is defined as follows:

a) $\alpha \oplus_{\min} \beta = \langle \sqrt{\mu_\alpha^2 + \mu_\beta^2 - \mu_\alpha^2 \mu_\beta^2}, \nu_1 \nu_2; \min(r_\alpha, r_\beta)\rangle$

b) $\alpha \oplus_{\max} \beta = \langle \sqrt{\mu_\alpha^2 + \mu_\beta^2 - \mu_\alpha^2 \mu_\beta^2}, \nu_1 \nu_2; \max(r_\alpha, r_\beta)\rangle$

c) $\alpha \otimes_{\min} \beta = \langle \mu_\alpha \mu_\beta, \sqrt{\nu_\alpha^2 + \nu_\beta^2 - \nu_\alpha^2 \nu_\beta^2}; \min(r_\alpha, r_\beta)\rangle$

d) $\alpha \otimes_{\max} \beta = \langle \mu_\alpha \mu_\beta, \sqrt{\nu_\alpha^2 + \nu_\beta^2 - \nu_\alpha^2 \nu_\beta^2}; \max(r_\alpha, r_\beta)\rangle$

Algebraic operations among C-PFVs in Definition 12 can be extended by using general $t$-norms and $t$-conorms.

**Definition 13** Let $\alpha = \langle \mu_\alpha, \nu_\alpha; r_\alpha\rangle$ and $\beta = \langle \mu_\beta, \nu_\beta; r_\beta\rangle$ be two C-PFVs. Assume that $T, S$ are dual $t$-norm and $t$-conorm with respect to the Pythagorean fuzzy complement $N(a) = \sqrt{1 - a^2}$, respectively and $Q$ is a $t$-norm or a $t$-conorm. General algebraic operations

*among C-PFVs is defined as follows:*

**a)** $\alpha \oplus_Q \beta = \langle S(\mu_\alpha, \mu_\beta), T(\mu_\alpha, \mu_\beta); Q(r_\alpha, r_\beta) \rangle$,

**b)** $\alpha \otimes_Q \beta = \langle T(\mu_\alpha, \mu_\beta), S(\mu_\alpha, \mu_\beta); Q(r_\alpha, r_\beta) \rangle$.

It is clear that with a particular choose of $S, T$ and $Q$ the operations given in Definition 12 are obtained from the operations defined in Definition 13.

We now show that the sum and the product of two C-PFVs are also C-PFVs with the following proposition.

**Proposition 2** *Let $\alpha$ and $\beta$ be two C-PFVs. Assume that $T, S$ are dual t-norm and t-conorm with respect to Pythagorean fuzzy complement $N(a) = \sqrt{1-a^2}$, respectively and $Q$ is a t-norm or a t-conorm. Then $\alpha \oplus_Q \beta$ and $\alpha \otimes_Q \beta$ are also C-PFVs.*

*Proof.* We know that the dual $t$-conorm $S$ with respect to the Pythagorean fuzzy complement $N$ is $S(x,y) = \sqrt{1 - T^2(\sqrt{1-x^2}, \sqrt{1-y^2})}$ from Remark 1. Since $\mu_\alpha \leq \sqrt{1 - v_\alpha^2}$ and $T$ is increasing, it is obtained that

$$T^2(\mu_\alpha, \mu_\beta) + S^2(v_\alpha, v_\beta) = T^2(\mu_\alpha, \mu_\beta) + (\sqrt{1 - T^2(\sqrt{1-v_\alpha^2}, \sqrt{1-v_\beta^2})})^2$$

$$= T^2(\mu_\alpha, \mu_\beta) + 1 - T^2(\sqrt{1-v_\alpha^2}, \sqrt{1-v_\beta^2})$$

$$\leq T^2(\sqrt{1-v_\alpha^2}, \sqrt{1-v_\beta^2}) + 1 - T^2(\sqrt{1-v_\alpha^2}, \sqrt{1-v_\beta^2})$$

$$= 1.$$

Moreover, since the domain of $Q$ is the unit closed interval we conclude that $\alpha \oplus_Q \beta$ is a C-PFV. Similarly, it can be shown that $\alpha \otimes_Q \beta$ is a also C-PFV.

Klement and Mesiar [19] showed that continuous Archimedean $t$-norms and $t$-conorms can be expressed with their additive generators. Thus, some algebraic operations among C-PFVs can be defined using additive generators of strict Archimedean $t$-norms and $t$-conorms.

**Definition 14** *Let $\alpha = \langle \mu_\alpha, v_\alpha; r_\alpha \rangle$ and $\beta = \langle \mu_\beta, v_\beta; r_\beta \rangle$ be two C-PFVs and let $\lambda > 0$. Assume that $g: [0,1] \to [0,\infty]$ is the additive generator of a continuous Archimedean t-norm and $h(t) = g(\sqrt{1-t^2})$ and $q: [0,1] \to [0,\infty]$ is the additive generator of a continuous Archimedean t-norm or t-conorm. Some algebraic operations among C-PFVs is defined as follows:*

**a)** $\alpha \oplus_q \beta = \langle h^{-1}(h(\mu_\alpha) + h(\mu_\beta)), g^{-1}(g(\mu_\alpha) + g(\mu_\beta)); q^{-1}(q(r_\alpha) + q(r_\beta)) \rangle$,

**b)** $\alpha \otimes_q \beta = \langle g^{-1}(g(\mu_\alpha) + g(\mu_\beta)), h^{-1}(h(v_\alpha) + h(v_\beta)); q^{-1}(q(r_\alpha) + q(r_\beta)) \rangle$,

**c)** $\lambda_q \alpha = \langle h^{-1}(\lambda h(\mu_\alpha)), g^{-1}(\lambda g(v_\alpha)); q^{-1}(\lambda q(r_\alpha)) \rangle$,

**d)** $\alpha^{\lambda_q} = \langle g^{-1}(\lambda g(\mu_\alpha)), h^{-1}(\lambda h(v_\alpha)); q^{-1}(\lambda q(r_\alpha)) \rangle$.

The following proposition confirms that multiplication by constant and power of C-PFVs are also C-PFVs.

**Proposition 3** Let $\alpha = \langle \mu_\alpha, \nu_\alpha; r_\alpha \rangle$ and $\beta = \langle \mu_\beta, \nu_\beta; r_\beta \rangle$ be two C-PFVs and let $\lambda > 0$. Assume that $g: [0,1] \to [0, \infty]$ is the additive generator of a continuous Archimedean t-norm and $h(t) = g(\sqrt{1-t^2})$ and $q: [0,1] \to [0, \infty]$ is the additive generator of a continuous Archimedean t-norm or t-conorm. Then $\alpha \oplus_q \beta$, $\alpha \otimes_q \beta$, $\lambda_q \alpha$ and $\alpha^{\lambda_q}$ are C-PFVs.

*Proof.* It is clear from Proposition 2 that $\alpha \oplus_q \beta$ and $\alpha \otimes_q \beta$ are C-PFVs. We know that $h^{-1}(t) = \sqrt{1 - [g^{-1}(t)]^2}$ and $g(t) = h(\sqrt{1-t^2})$. Since $\mu_\alpha \leq \sqrt{1-\nu_\alpha^2}$ and $h, h^{-1}$ are increasing, we have

$$0 \leq [h^{-1}(\lambda h(\mu_\alpha))]^2 + [g^{-1}(\lambda g(\nu_\alpha))]^2$$
$$\leq [h^{-1}(\lambda h(\sqrt{1-\nu_\alpha^2}))]^2 + [g^{-1}(\lambda g(\nu_\alpha))]^2$$
$$= 1 - [g^{-1}(\lambda h(\sqrt{1-\nu_\alpha^2}))]^2 + [g^{-1}(\lambda g(\nu_\alpha))]^2$$
$$= 1 - [g^{-1}(\lambda g(\nu_\alpha))]^2 + [g^{-1}(\lambda g(\nu_\alpha))]^2$$
$$= 1.$$

Moreover, since the domain of $q^{-1}$ is the unit closed interval we conclude that $\lambda_q \alpha$ is a C-PFV. In a similar way, it can be shown that $\alpha^{\lambda_q}$ is also a C-PFV.

**Example 4** Let $g, h, q, p: [0,1] \to [0, \infty]$ defined by $g(t) = -\log t^2$, $h(t) = -\log(1-t^2)$, $q(t) = -\log t^2$ and $p(t) = -\log(1-t^2)$ and $\lambda > 0$. Then we obtain the algebraic operators

a) $\alpha \oplus_q \beta = \langle \sqrt{\mu_\alpha^2 + \mu_\beta^2 - \mu_\alpha^2 \mu_\beta^2}, \nu_\alpha \nu_\beta; r_\alpha r_\beta \rangle$

b) $\alpha \oplus_p \beta = \langle \sqrt{\mu_\alpha^2 + \mu_\beta^2 - \mu_\alpha^2 \mu_\beta^2}, \nu_\alpha \nu_\beta; \sqrt{r_\alpha^2 + r_\beta^2 - r_\alpha^2 r_\beta^2} \rangle$

c) $\alpha \otimes_q \beta = \langle \mu_\alpha \mu_\beta, \sqrt{\nu_\alpha^2 + \nu_\beta^2 - \nu_\alpha^2 \nu_\beta^2}; r_\alpha r_\beta \rangle$

d) $\alpha \otimes_p \beta = \langle \mu_\alpha \mu_\beta, \sqrt{\nu_\alpha^2 + \nu_\beta^2 - \nu_\alpha^2 \nu_\beta^2}; \sqrt{r_\alpha^2 + r_\beta^2 - r_\alpha^2 r_\beta^2} \rangle$

e) $\lambda_q \alpha = \langle \sqrt{1 - (1-\mu_\alpha^2)^\lambda}, \nu_\alpha^\lambda; r_\alpha^\lambda \rangle$

f) $\lambda_p \alpha = \langle \sqrt{1 - (1-\mu_\alpha^2)^\lambda}, \nu_\alpha^\lambda; \sqrt{1 - (1-r_\alpha^2)^\lambda} \rangle$

g) $\alpha^{\lambda_q} = \langle \mu_\alpha^\lambda, \sqrt{1 - (1-\nu_\alpha^2)^\lambda}; r_\alpha^\lambda \rangle$

h) $\alpha^{\lambda_p} = \langle \mu_\alpha^\lambda, \sqrt{1 - (1-\nu_\alpha^2)^\lambda}; \sqrt{1 - (1-r_\alpha^2)^\lambda} \rangle$

Following theorem gives some basic properties of algebraic operations.

**Theorem 3** Let $\alpha = \langle \mu_\alpha, \nu_\alpha; r_\alpha \rangle$, $\beta = \langle \mu_\beta, \nu_\beta; r_\beta \rangle$ and $\theta = \langle \mu_\theta, \nu_\theta; r_\theta \rangle$ be C-PFVs and let $\lambda, \gamma > 0$. Assume that $g: [0,1] \to [0, \infty]$ is the additive generator of a continuous Archimedean t-norm and $h(t) = g(\sqrt{1-t^2})$ and $q: [0,1] \to [0, \infty]$ is the additive generator of a continuous Archimedean t-norm or t-conorm. We have

i) $\alpha \oplus_q \beta = \beta \oplus_q \alpha$

ii) $\alpha \otimes_q \beta = \beta \otimes_q \alpha$
iii) $(\alpha \oplus_q \beta) \oplus_q \theta = \alpha \oplus_q (\beta \oplus_q \theta)$
iv) $(\alpha \otimes_q \beta) \otimes_q \theta = \alpha \otimes_q (\beta \otimes_q \theta)$
v) $\lambda_q(\alpha \oplus_q \beta) = \lambda_q \alpha \oplus_q \lambda_q \beta$
vi) $(\lambda_q + \gamma_q)\alpha = \lambda_q \alpha \oplus_q \gamma_q \alpha$
vii) $(\alpha \otimes_q \beta)^{\lambda_q} = \alpha^{\lambda_q} \otimes_q \beta^{\lambda_q}$
viii) $\alpha^{\lambda_q} \otimes_q \alpha^{\gamma_q} = \alpha^{\lambda_q + \gamma_q}$

*Proof.* (i) and (ii) are trivial.

**iii)** We obtain

$(\alpha \oplus_q \beta) \oplus_q \theta = \langle h^{-1}(h(\mu_\alpha) + h(\mu_\beta)), g^{-1}(g(\mu_\alpha) + g(\mu_\beta)); q^{-1}(q(r_\alpha) + q(r_\beta))\rangle \oplus_q \langle \mu_\theta, \nu_\theta; r_\theta\rangle$

$= \langle h^{-1}(h(h^{-1}(h(\mu_\alpha) + h(\mu_\beta)) + h(\mu_\theta)), g^{-1}(g(g^{-1}(g(\mu_\alpha) + g(\mu_\beta)) + g(\mu_\theta));$

$q^{-1}(q(q^{-1}(q(r_\alpha) + q(r_\beta)) + q(r_\theta)))\rangle$

$= \langle h^{-1}(h(\mu_\alpha) + h(\mu_\beta) + h(\mu_\theta)), g^{-1}(g(\mu_\alpha) + g(\mu_\beta) + g(\mu_\theta));$

$q^{-1}(q(r_\alpha) + q(r_\beta) + q(r_\theta))\rangle$

$= \langle h^{-1}(h(\mu_\alpha) + h(h^{-1}(h(\mu_\beta) + h(\mu_\theta)))), g^{-1}(g(\mu_\alpha) + g(g^{-1}(g(\mu_\beta) + g(\mu_\theta))));$

$q^{-1}(q(r_\alpha) + q(q^{-1}(q(r_\beta) + q(r_\theta))))\rangle$

$= \langle \mu_\alpha, \nu_\alpha; r_\alpha\rangle \oplus_q \langle h^{-1}(h(\mu_\beta) + h(\mu_\theta)), g^{-1}(g(\mu_\beta) + g(\mu_\theta)); q^{-1}(q(r_\beta) + q(r_\theta))\rangle$

$= \alpha \oplus_q (\beta \oplus_q \theta)$.

**iv)** It is obtained that

$(\alpha \otimes_q \beta) \otimes_q \theta = \langle g^{-1}(g(\mu_\alpha) + g(\mu_\beta)), h^{-1}(h(\mu_\alpha) + h(\mu_\beta)); q^{-1}(q(r_\alpha) + q(r_\beta))\rangle \otimes_q \langle \mu_\theta, \nu_\theta; r_\theta\rangle$

$= \langle g^{-1}(g(g^{-1}(g(\mu_\alpha) + g(\mu_\beta)) + g(\mu_\theta)), h^{-1}(h(h^{-1}(h(\mu_\alpha) + h(\mu_\beta)) + h(\mu_\theta));$

$q^{-1}(q(q^{-1}(q(r_\alpha) + q(r_\beta)) + q(r_\theta)))\rangle$

$= \langle g^{-1}(g(\mu_\alpha) + g(\mu_\beta) + g(\mu_\theta)), h^{-1}(h(\mu_\alpha) + h(\mu_\beta) + h(\mu_\theta));$

$q^{-1}(q(r_\alpha) + q(r_\beta) + q(r_\theta))\rangle$

$= \langle g^{-1}(g(\mu_\alpha) + g(g^{-1}(g(\mu_\beta) + g(\mu_\theta)))), h^{-1}(h(\mu_\alpha) + h(h^{-1}(h(\mu_\beta) + h(\mu_\theta))));$

$q^{-1}(q(r_\alpha) + q(q^{-1}(q(r_\beta) + q(r_\theta))))\rangle$

$= \langle \mu_\alpha, \nu_\alpha; r_\alpha\rangle \otimes_q \langle g^{-1}(g(\mu_\beta) + g(\mu_\theta)), h^{-1}(h(\mu_\beta) + h(\mu_\theta)); q^{-1}(q(r_\beta) + q(r_\theta))\rangle$

$= \alpha \otimes_q (\beta \otimes_q \theta)$.

**v)** We get

$\lambda_q(\alpha \oplus_q \beta) = \lambda_q\langle h^{-1}(h(\mu_\alpha) + h(\mu_\beta)), g^{-1}(g(\mu_\alpha) + g(\mu_\beta)); q^{-1}(q(r_\alpha) + q(r_\beta))\rangle$

$= \langle h^{-1}(\lambda h(h^{-1}(h(\mu_\alpha) + h(\mu_\beta)))), g^{-1}(\lambda g(g^{-1}(g(\mu_\alpha) + g(\mu_\beta))));$

$q^{-1}(\lambda q(q^{-1}(q(r_\alpha) + q(r_\beta))))\rangle$

$$= \langle h^{-1}(\lambda h(\mu_\alpha) + \lambda h(\mu_\beta)), g^{-1}(\lambda g(\mu_\alpha) + \lambda g(\mu_\beta)); q^{-1}(\lambda q(r_\alpha) + \lambda q(r_\beta))\rangle$$
$$= \langle h^{-1}(h(h^{-1}(\lambda h(\mu_\alpha))) + h(h^{-1}(\lambda h(\mu_\beta)))), g^{-1}(g(g^{-1}(\lambda g(\mu_\alpha))) + g(g^{-1}(\lambda g(\mu_\beta))));$$
$$q^{-1}(q(q^{-1}(\lambda q(r_\alpha))) + q(q^{-1}(\lambda q(r_\beta))))\rangle$$
$$= \langle h^{-1}(h(\mu_{\lambda\alpha}) + h(\mu_{\lambda\beta})), g^{-1}(g(\mu_{\lambda\alpha}) + g(\mu_{\lambda\beta})); q^{-1}(q(r_{\lambda\alpha}) + \lambda q(s_{\lambda\beta}))\rangle$$
$$= \lambda_q \alpha \oplus_q \lambda_q \beta.$$

**vi)** It is clear that
$$(\lambda_q + \gamma_q)\alpha = \langle h^{-1}((\lambda+\gamma)h(\mu_\alpha)), g^{-1}((\lambda+\gamma)g(\nu_\alpha)); q^{-1}((\lambda+\gamma)q(r_\alpha))\rangle$$
$$= \langle h^{-1}(\lambda h(\mu_\alpha) + \gamma h(\mu_\alpha)), g^{-1}(\lambda g(\nu_\alpha) + \gamma g(\nu_\alpha)); q^{-1}(\lambda q(r_\alpha) + \gamma q(r_\alpha))\rangle$$
$$= \langle h^{-1}(h(h^{-1}(\lambda h(\mu_\alpha))) + h(h^{-1}(\gamma h(\mu_\alpha)))), g^{-1}(g(g^{-1}(\lambda g(\nu_\alpha))) + g(g^{-1}(\gamma g(\nu_\alpha))));$$
$$q^{-1}(q(q^{-1}(\lambda q(r_\alpha))) + q(q^{-1}(\gamma q(r_\alpha))))\rangle$$
$$= \langle h^{-1}(h(\mu_{\lambda_q\alpha}) + h(\mu_{\gamma_q\alpha})), g^{-1}(g(\nu_{\lambda_q\alpha}) + g(\nu_{\gamma_q\alpha})); q^{-1}(q(r_{\lambda_q\alpha}) + q(r_{\gamma_q\alpha}))\rangle$$
$$= \lambda_q \alpha \oplus_q \gamma_q \alpha.$$

**vii)** We have
$$(\alpha \otimes_q \beta)^{\lambda_q} = \langle g^{-1}(\lambda g(\mu_{\alpha\otimes_q\beta})), h^{-1}(\lambda h(\nu_{\alpha\otimes_q\beta})); q^{-1}(\lambda q(r_{\alpha\otimes_q\beta}))\rangle$$
$$= \langle g^{-1}(\lambda g(g^{-1}(g(\mu_\alpha) + g(\mu_\beta)))), h^{-1}(\lambda h(h^{-1}(h(\nu_\alpha) + h(\nu_\beta))));$$
$$q^{-1}(\lambda q(q^{-1}(q(r_\alpha) + q(r_\beta))))\rangle$$
$$= \langle g^{-1}(\lambda g(\mu_\alpha) + \lambda g(\mu_\beta)), h^{-1}(\lambda h(\nu_\alpha) + \lambda h(\nu_\beta)); q^{-1}(\lambda q(r_\alpha) + \lambda q(r_\beta))\rangle$$
$$= \langle g^{-1}(g(g^{-1}(\lambda g(\mu_\alpha))) + g(g^{-1}(\lambda g(\mu_\beta)))), h^{-1}(h(h^{-1}(\lambda h(\nu_\alpha))) + h(h^{-1}(\lambda h(\nu_\beta))));$$
$$q^{-1}(q(q^{-1}(\lambda q(r_\alpha))) + q(q^{-1}(\lambda q(r_\beta))))\rangle$$
$$= \langle g^{-1}(g(\mu_{\alpha^{\lambda_q}}) + g(\mu_{\beta^{\lambda_q}})), h^{-1}(h(\nu_{\alpha^{\lambda_q}}) + h(\nu_{\beta^{\lambda_q}})); q^{-1}(q(r_{\alpha^{\lambda_q}}) + q(r_{\beta^{\lambda_q}}))\rangle$$
$$= \alpha^{\lambda_q} \otimes_q \beta^{\lambda_q}.$$

**viii)** We have
$$\alpha_r^{\lambda_q+\gamma_q} = \langle g^{-1}((\lambda+\gamma)g(\mu_\alpha)), h^{-1}((\lambda+\gamma)h(\nu_\alpha)); q^{-1}((\lambda+\gamma)q(r_\alpha))\rangle$$
$$= \langle g^{-1}(\lambda g(\mu_\alpha) + \gamma g(\mu_\alpha)), h^{-1}(\lambda h(\nu_\alpha) + \gamma h(\nu_\alpha)); q^{-1}(\lambda q(r_\alpha) + \gamma q(r_\alpha))\rangle$$
$$= \langle g^{-1}(g(g^{-1}(\lambda g(\mu_\alpha))) + g(g^{-1}(\gamma g(\mu_\alpha)))), h^{-1}(h(h^{-1}(\lambda h(\nu_\alpha))) + h(h^{-1}(\gamma h(\nu_\alpha))));$$
$$q^{-1}(q(q^{-1}(\lambda q(r_\alpha))) + q(q^{-1}(\gamma q(r_\alpha))))\rangle$$
$$= \langle g^{-1}(g(\mu_{\alpha^\lambda}) + g(\mu_{\alpha^\gamma})), h^{-1}(h(\nu_{\alpha^\lambda}) + h(\nu_{\alpha^\gamma})); q^{-1}(q(r_{\alpha^\lambda}) + q(r_{\alpha^\gamma}))\rangle$$
$$= \alpha_r^{\lambda_q} \otimes_q \alpha_r^{\gamma_q}.$$

## 4 Aggregation Operators For C-PFVs

Aggregation operators (see e.g. [5, 12, 17]) have an important role while transforming input values represented by fuzzy values to a single output value. In this section, we introduce a weighted arithmetic aggregation operator and a weighted geometric aggregation operator for C-PFVs by using algebraic operations given in Section 3.

### 4.1 Weighted Arithmetic Aggregation Operators

**Definition 15** Let $\{\alpha_i = \langle \mu_{\alpha_i}, \nu_{\alpha_i}; r_{\alpha_i} \rangle : i = 1, \ldots, n\}$ be a collection of C-PFVs. Assume that $g:[0,1] \to [0,\infty]$ is the additive generator of a continuous Archimedean t-norm and $h(t) = g(\sqrt{1-t^2})$ and $q:[0,1] \to [0,\infty]$ is the additive generator of a continuous Archimedean t-norm or a t-conorm. Then a weighted arithmetic aggregation operator $CPWA_q$ is defined by

$$CPWA_q(\alpha_1, \ldots, \alpha_n) := (q) \bigoplus_{i=1}^{n} w_{i_q} \alpha_i$$

where $0 \leq w_i \leq 1$ for any $i = 1, \ldots, n$ with $\sum_{i=1}^{n} w_i = 1$.

**Theorem 4** Let $\{\alpha_i = \langle \mu_{\alpha_i}, \nu_{\alpha_i}; r_i \rangle : i = 1, \ldots, n\}$ be a collection of C-PFVs. Assume that $g:[0,1] \to [0,\infty]$ is the additive generator of a continuous Archimedean t-norm and $h(t) = g(\sqrt{1-t^2})$ and $q:[0,1] \to [0,\infty]$ is the additive generator of a continuous Archimedean t-norm or a t-conorm. Then $CPWA_q(\alpha_1, \ldots, \alpha_n)$ is a C-PFV and we have

$CPWA_q(\alpha_1, \ldots, \alpha_n) =$
$\langle h^{-1}(\sum_{i=1}^{n} w_i h(\mu_{\alpha_i})), g^{-1}(\sum_{i=1}^{n} w_i g(\nu_{\alpha_i})); q^{-1}(\sum_{i=1}^{n} w_i q(r_{\alpha_i})) \rangle$
where $0 \leq w_i \leq 1$ for any $i = 1, \ldots, n$ with $\sum_{i=1}^{n} w_i = 1$.

*Proof.* It is seen from Proposition 3 that $CPWA_q(\alpha_1, \ldots, \alpha_n)$ is a C-PFV. By utilizing mathematical induction it can be seen that the second part is also true. If $n = 2$, we have

$CPWA_q(\alpha_1, \alpha_2) = w_{1_q} \alpha_1 \oplus_q w_{2_q} \alpha_2$
$= \langle h^{-1}(h(\mu_{w_{1_q}\alpha_1}) + h(\mu_{w_{2_q}\alpha_2})), g^{-1}(g(\nu_{w_{1_q}\alpha_1}) + g(\nu_{w_{2_q}\alpha_2}));$
$q^{-1}(q(r_{w_{1_q}\alpha_1}) + q(r_{w_{2_q}\alpha_2})) \rangle$
$= \langle h^{-1}(h(h^{-1}(w_1 h(\mu_{\alpha_1}))) + h(h^{-1}(w_1 h(\mu_{\alpha_1})))),$
$g^{-1}(g(g^{-1}(w_1 g(\nu_{\alpha_1}))) + g(g^{-1}(w_2 g(\nu_{\alpha_2}))));$
$q^{-1}(q(q^{-1}(w_1 q(r_{\alpha_1}))) + q(q^{-1}(w_2 q(r_{\alpha_2})))) \rangle$
$= \langle h^{-1}(w_1 h(\mu_{\alpha_1}) + w_2 h(\mu_{\alpha_2})), g^{-1}(w_1 g(\nu_{\alpha_1}) + w_2 g(\nu_{\alpha_2}));$
$q^{-1}(w_1 q(r_{\alpha_1}) + w_2 g(r_{\alpha_2})) \rangle$
$= \langle h^{-1}(\sum_{j=1}^{2} w_j h(\mu_{\alpha_j})), g^{-1}(\sum_{j=1}^{2} w_j g(\nu_{\alpha_j})); q^{-1}(\sum_{j=1}^{2} w_j q(r_{\alpha_j})) \rangle.$

Now assume that the expression
$A_{n-1} = CPWA_q(\alpha_1, \ldots, \alpha_{n-1}) =$
$\langle h^{-1}(\sum_{j=1}^{n-1} w_j h(\mu_{\alpha_j})), g^{-1}(\sum_{j=1}^{n-1} w_j g(\nu_{\alpha_j})); q^{-1}(\sum_{j=1}^{n-1} w_j q(r_{\alpha_j})) \rangle.$
is valid. Then we have

$CPWA_q(\alpha_1, \ldots, \alpha_n) = A_{n-1} \oplus_q w_{n_q} \alpha_n$
$= \langle h^{-1}(\sum_{j=1}^{n-1} w_j h(\mu_{\alpha_j})), g^{-1}(\sum_{j=1}^{n-1} w_j g(\nu_{\alpha_j})); q^{-1}(\sum_{j=1}^{n-1} w_j q(r_{\alpha_j})) \rangle \oplus_q$
$\langle h^{-1}(w_n(\mu_{\alpha_n})), g^{-1}(w_n g(\nu_{\alpha_n})); q - 1(w_n q(r_{\alpha_n})) \rangle$
$= \langle h^{-1}(h(h^{-1}(\sum_{j=1}^{n-1} w_j h(\mu_{\alpha_j}))) + h(h^{-1}(w_n h(\mu_{\alpha_n})))),$
$g^{-1}(g(g^{-1}(\sum_{j=1}^{n-1} w_j g(\nu_{\alpha_j}))) + g(g^{-1}(w_n g(\nu_{\alpha_n}))));$
$q^{-1}(q(q^{-1}(\sum_{j=1}^{n-1} w_j q(\nu_{\alpha_j}))) + q(q^{-1}(w_n q(\nu_{\alpha_n})))) \rangle$
$= \langle h^{-1}(\sum_{j=1}^{n-1} w_j h(\mu_{\alpha_j}) + w_n h(\mu_{\alpha_n})), g^{-1}(\sum_{j=1}^{n-1} w_j g(\nu_{\alpha_j}) + w_n g(\nu_{\alpha_n}));$
$q^{-1}(\sum_{j=1}^{n-1} w_j q(r_{\alpha_j}) + w_n q(r_{\alpha_n})) \rangle$

$$= \langle h^{-1}(\textstyle\sum_{i=1}^{n} w_i h(\mu_{\alpha_i})), g^{-1}(\sum_{i=1}^{n} w_i g(v_{\alpha_i})); q^{-1}(\sum_{i=1}^{n} w_i q(r_{\alpha_i}))\rangle.$$

Thus, the proof is completed.

**Remark 4** *Let $g, h, q, p: [0,1] \to [0, \infty]$ be functions defined by $g(t) = -\log t^2$, $h(t) = -\log(1 - t^2)$, $q(t) = -\log t^2$ and $p(t) = -\log(1 - t^2)$. Then we obtain Algebraic weighted arithmetic aggregation operators as particular cases of the aggregation operators given in Definition 15 as follows:*

$$CPWA_q^A(\alpha_1, \ldots, \alpha_n) = \langle \sqrt{1 - \textstyle\prod_{i=1}^{n}(1 - \mu_{\alpha_i}^2)^{w_i}}, \prod_{i=1}^{n} v_{\alpha_i}^{w_i}; \prod_{i=1}^{n} r_{\alpha_i}^{w_i}\rangle$$

and

$$CPWA_p^A(\alpha_1, \ldots, \alpha_n) =$$
$$\langle \sqrt{1 - \textstyle\prod_{i=1}^{n}(1 - \mu_{\alpha_i}^2)^{w_i}}, \prod_{i=1}^{n} v_{\alpha_i}^{w_i}; \sqrt{1 - \prod_{i=1}^{n}(1 - r_{\alpha_i}^2)^{w_i}}\rangle.$$

### 4.2 Weighted Geometric Aggregation Operators

**Definition 16** *Let $\{\alpha_i = \langle \mu_{\alpha_i}, v_{\alpha_i}; r_{\alpha_i}\rangle : i = 1, \ldots, n\}$ be a collection of C-PFVs. Assume that $g: [0,1] \to [0, \infty]$ is the additive generator of a continuous Archimedean t-norm and $h(t) = g(\sqrt{1 - t^2})$ and $q: [0,1] \to [0, \infty]$ is the additive generator of a continuous Archimedean t-norm or a t-conorm. Then a weighted geometric aggregation operator $CPWG_q$ is defined by*

$$CPWG_q(\alpha_1, \ldots, \alpha_n) := (q) \bigotimes_{i=1}^{n} w_i \alpha_i$$

*where $0 \leq w_i \leq 1$ for any $i = 1, \ldots, n$ with $\sum_{i=1}^{n} w_i = 1$.*

**Theorem 5** *Let $\{\alpha_i = \langle \mu_{\alpha_i}, v_{\alpha_i}; r_{\alpha_i}\rangle : i = 1, \ldots, n\}$ be a collection of C-PFVs. Assume that $g: [0,1] \to [0, \infty]$ is the additive generator of a continuous Archimedean t-norm and $h(t) = g(\sqrt{1 - t^2})$ and $q: [0,1] \to [0, \infty]$ is the additive generator of a continuous Archimedean t-norm or a t-conorm. Then*
$$CPWG_q(\alpha_1, \ldots, \alpha_n) =$$
$$\langle g^{-1}(\textstyle\sum_{i=1}^{n} w_i g(\mu_{\alpha_i})), h^{-1}(\sum_{i=1}^{n} w_i h(v_{\alpha_i})); q^{-1}(\sum_{i=1}^{n} w_i q(r_{\alpha_i}))\rangle$$
*where $0 \leq w_i \leq 1$ for any $i = 1, \ldots, n$ with $\sum_{i=1}^{n} w_i = 1$.*

*Proof.* It can be proved similar to Theorem 4.

**Remark 5** *Let $g, h, q, p: [0,1] \to [0, \infty]$ be functions defined by $g(t) = -\log t^2$, $h(t) = -\log(1 - t^2)$, $q(t) = -\log t^2$ and $p(t) = -\log(1 - t^2)$. Then we obtain Algebraic weighted geometric aggregation operators as particular cases of the aggregation operators given in Definition 16 as follows:*

$$CPWG_q^A(\alpha_1, \ldots, \alpha_n) = \langle \textstyle\prod_{i=1}^{n} \mu_{\alpha_i}^{w_i}, \sqrt{1 - \prod_{i=1}^{n}(1 - v_{\alpha_i}^2)^{w_i}}; \prod_{i=1}^{n} r_{\alpha_i}^{w_i}\rangle.$$

and

$$CPWG_p^A(\alpha_1, \ldots, \alpha_n) =$$
$$\langle \prod_{i=1}^n \mu_{\alpha_i}^{w_i}, \sqrt{1 - \prod_{i=1}^n (1-v_{\alpha_i}^2)^{w_i}}; \sqrt{1 - \prod_{i=1}^n (1-r_{\alpha_i}^2)^{w_i}} \rangle.$$

## 5  An Application of C-PFVs to a MCDM Problem

In this section, we define a similarity measure for C-PFVs. Then using this similarity measure and the proposed aggregation operators we propose a MCDM method in circular Pythagorean fuzzy environment. Then we solve a real world decision problem from the literature [39] that deals with selecting the best photovoltaic cell by utilizing the proposed method.

### 5.1  A Similarity Measure for C-PFVs

Similarity measures have an important role in the determination of the degree of similarity between two objects. Particularly, similarity measures for PFVs or PFSs have been investigated and developed by researchers since they are important tools for decision making, image processing, pattern recognition, classification and some other real life areas. Motivating from the cosine similarity measure for PFVs defined in [30], we give the following similarity measure for C-PFVs.

**Definition 17**  Let $\alpha = \langle \mu_\alpha, v_\alpha; r_\alpha \rangle$ and $\beta = \langle \mu_\beta, v_\beta; r_\beta \rangle$ be two C-PFVs. The cosine similarity measure CSM is defined by

$$CSM(\alpha, \beta) = \frac{1}{2}\left(\frac{\mu_\alpha^2 \mu_\beta^2 + v_\alpha^2 v_\beta^2}{\sqrt{\mu_\alpha^4 + v_\alpha^4}\sqrt{\mu_\beta^4 + v_\beta^4}} + 1 - |r_\alpha - r_\beta|\right).$$

**Theorem 6** Let $\alpha = \langle \mu_\alpha, v_\alpha; r_\alpha \rangle$ and $\beta = \langle \mu_\beta, v_\beta; r_\beta \rangle$ be two C-PFVs. The cosine similarity measure $CSM$ based on radius satisfies the following properties:
  i) $0 \leq CSM(\alpha, \beta) \leq 1$
  ii) $CSM(\alpha, \beta) = CSM(\beta, \alpha)$
  iii) If $\alpha = \beta$, then $CSM(\alpha, \beta) = 1$.

*Proof.* **i)** It is clear that $0 \leq 1 - |r_\alpha - r_\beta| \leq 1$. On the other hand the expression

$$\frac{\mu_\alpha^2 \mu_\beta^2 + v_\alpha^2 v_\beta^2}{\sqrt{\mu_\alpha^4 + v_\alpha^4}\sqrt{\mu_\beta^4 + v_\beta^4}}$$

is the cosine value of a certain angle in $\left[0, \frac{\pi}{2}\right]$ specified by $\alpha$ and $\beta$. So it is in the unit closed interval. Thus we have $0 \leq CSM(\alpha, \beta) \leq 1$.

The proof of (ii) and (iii) is trivial from the definition of $CSM$.

### 5.2  A MCDM method

In this sub-section a MCDM method is proposed in the circular Pythagorean fuzzy environment. The proposed method is applied to a MCDM problem adapted from the literature [39] to show the efficiency of this method in next sub-section. We can present steps of the proposed method as follows:

**Step 1:** Consider a set of $k$ alternatives as $A = \{A_1, \ldots, A_k\}$ evaluated by an expert with respect to a set of $j$ criteria as $C = \{x_1, \ldots, x_j\}$.

**Step 2:** The expert expresses the evaluation results of alternatives as C-PFVs according to each criterion and determines the weight vector.

**Step 3:** If there exists a cost criterion, then the complement operation is taken to the values of this criterion.

**Step 4:** Using proposed weighted aggregation operators, evaluation results expressed as C-PFVs for each alternatives are transformed to a value expressed as C-PFVs.

**Step 5:** The cosine similarity measure $CSM$ between aggregated value of each alternative and positive ideal alternative $\langle 1,0; 1 \rangle$ are calculated.

**Step 6:** Alternatives are ranked so that the maximum similarity value is the best alternative.

## 5.3 Evaluation of the problem of selecting photovoltaic cells

Due to the scarcity of non-renewable energies and their harmful effects on the environment, the importance of renewable energy sources has increased gradually for supplying plentiful and clean energy. One of the current renewable energy sources is photovoltaic cell, which have almost no negative effects on the environment and is enormously productive. A photovoltaic cell, also known as a solar cell, is an energy generating device that converts solar energy into electricity by the photovoltaic effect, which is a conversion discovered by Becquerel [4]. Choosing the best photovoltaic cell has an important role to increase production, to reduce costs and to confer high maturity and reliability. There are many types of photovoltaic cells. The aim of this section is to solve a MCDM problem adapted from the literature [39] about selecting the best photovoltaic cell. In [27], the photovoltaic cells forms the alternatives of MCDM problem and these alternatives are the following:

$A_1$: Photovoltaic cells with crystalline silicon (mono-crystalline and poly-crystalline),

$A_2$: Photovoltaic cells with inorganic thin layer (amorphous silicon),

$A_3$: Photovoltaic cells with inorganic thin layer (cadmium telluride/cadmium sulfide and copper indium gallium diselenide/cadmium sulfide),

$A_4$: Photovoltaic cells with advanced III–V thin layer with tracking systems for solar concentration, and

$A_5$: Photovoltaic cells with advanced, low cost, thin layers (organic and hybrid cells).

After viewing the photovoltaic cells determined as alternatives in the study, the criteria considered for the assessment of MCDM are the following: (1) $x_1$ (manufacturing cost), (2) $x_2$ (efficiency in energy conversion), (3) $x_3$ (market share), (4) $x_4$ (emissions of greenhouse gases generated during the manufacturing process), and (5) $x_5$ (energy payback time). It is noted that the criteria $x_2$ and $x_3$ are the benefit criteria, and others are the cost criteria. According to these five criteria, three experts specializing in photovoltaic systems and technologies evaluate these five available photovoltaic cells. The weight vector of the criteria determined by experts is $w = (0.2, 0.4, 0.1, 0.1, 0.2)$, and the weight vector of experts is fully unknown (see, [27]).

Now let us consider this problem with the method developed in the present paper. Steps 1-2 are already conducted. Table 1 is the decision matrix taken from [27].

| Experts | Alternatives | $C_1$ | $C_2$ | $C_3$ | $C_4$ | $C_5$ |
|---|---|---|---|---|---|---|
| $E_1$ | $A_1$ | ⟨0.8,0.4⟩ | ⟨0.8,0.6⟩ | ⟨0.6,0.7⟩ | ⟨0.8,0.3⟩ | ⟨0.6,0.5⟩ |
| | $A_2$ | ⟨0.5,0.7⟩ | ⟨0.9,0.2⟩ | ⟨0.8,0.5⟩ | ⟨0.6,0.3⟩ | ⟨0.5,0.6⟩ |
| | $A_3$ | ⟨0.4,0.3⟩ | ⟨0.3,0.7⟩ | ⟨0.7,0.4⟩ | ⟨0.4,0.6⟩ | ⟨0.5,0.4⟩ |
| | $A_4$ | ⟨0.6,0.6⟩ | ⟨0.7,0.5⟩ | ⟨0.7,0.2⟩ | ⟨0.6,0.4⟩ | ⟨0.7,0.3⟩ |
| | $A_5$ | ⟨0.7,0.5⟩ | ⟨0.6,0.4⟩ | ⟨0.9,0.3⟩ | ⟨0.7,0.6⟩ | ⟨0.7,0.1⟩ |
| $E_2$ | $A_1$ | ⟨0.9,0.3⟩ | ⟨0.7,0.6⟩ | ⟨0.5,0.8⟩ | ⟨0.6,0.3⟩ | ⟨0.6,0.3⟩ |
| | $A_2$ | ⟨0.4,0.7⟩ | ⟨0.9,0.2⟩ | ⟨0.8,0.1⟩ | ⟨0.5,0.3⟩ | ⟨0.5,0.3⟩ |
| | $A_3$ | ⟨0.6,0.3⟩ | ⟨0.7,0.7⟩ | ⟨0.7,0.6⟩ | ⟨0.4,0.4⟩ | ⟨0.3,0.4⟩ |
| | $A_4$ | ⟨0.8,0.4⟩ | ⟨0.7,0.5⟩ | ⟨0.6,0.2⟩ | ⟨0.7,0.4⟩ | ⟨0.7,0.4⟩ |
| | $A_5$ | ⟨0.7,0.2⟩ | ⟨0.8,0.2⟩ | ⟨0.8,0.4⟩ | ⟨0.6,0.6⟩ | ⟨0.6,0.6⟩ |
| $E_3$ | $A_1$ | ⟨0.8,0.6⟩ | ⟨0.7,0.6⟩ | ⟨0.5,0.8⟩ | ⟨0.5,0.5⟩ | ⟨0.6,0.1⟩ |
| | $A_2$ | ⟨0.5,0.6⟩ | ⟨0.9,0.2⟩ | ⟨0.8,0.1⟩ | ⟨0.5,0.3⟩ | ⟨0.4,0.3⟩ |
| | $A_3$ | ⟨0.7,0.4⟩ | ⟨0.7,0.5⟩ | ⟨0.6,0.1⟩ | ⟨0.9,0.2⟩ | ⟨0.5,0.6⟩ |
| | $A_4$ | ⟨0.9,0.2⟩ | ⟨0.5,0.6⟩ | ⟨0.6,0.2⟩ | ⟨0.6,0.1⟩ | ⟨0.7,0.4⟩ |
| | $A_5$ | ⟨0.6,0.1⟩ | ⟨0.8,0.2⟩ | ⟨0.9,0.2⟩ | ⟨0.5,0.6⟩ | ⟨0.6,0.4⟩ |

Table 1: Pythagorean fuzzy group decision matrix

**Step 3:** Since $x_1, x_4$ and $x_5$ are the cost criteria, we take the complement of these values. Thus we obtain Pythagorean fuzzy group normalized decision matrix shown in Table 2. As this decision matrix consists of PFVs we need to convert these values to C-PFVs. For this purpose we use Proposition 1. For example, according to the $x_1$ criterion of the $A_1$ alternative, the evaluation results of the experts are ⟨0.4,0.8⟩, ⟨0.3,0.9⟩, ⟨0.6,0.8⟩, respectively. From Proposition 1, it is seen that the arithmetic average of the evaluation results is ⟨0.45,0.83⟩ and the radius is 0.16. In this way, we attain aggregated Pythagorean fuzzy decision matrix given in Table 3 and maximum radius lengths based on decision matrix listed in Table 4. With C-PFVs the circular Pythagorean fuzzy decision matrix is shown in Table 5.

| Experts | Alternatives | $C_1$ | $C_2$ | $C_3$ | $C_4$ | $C_5$ |
|---|---|---|---|---|---|---|
| $E_1$ | $A_1$ | ⟨0.4,0.8⟩ | ⟨0.8,0.6⟩ | ⟨0.6,0.7⟩ | ⟨0.3,0.8⟩ | ⟨0.5,0.6⟩ |
| | $A_2$ | ⟨0.7,0.5⟩ | ⟨0.9,0.2⟩ | ⟨0.8,0.5⟩ | ⟨0.3,0.6⟩ | ⟨0.6,0.5⟩ |
| | $A_3$ | ⟨0.3,0.4⟩ | ⟨0.3,0.7⟩ | ⟨0.7,0.4⟩ | ⟨0.6,0.4⟩ | ⟨0.4,0.5⟩ |
| | $A_4$ | ⟨0.6,0.6⟩ | ⟨0.7,0.5⟩ | ⟨0.7,0.2⟩ | ⟨0.4,0.6⟩ | ⟨0.3,0.7⟩ |
| | $A_5$ | ⟨0.5,0.7⟩ | ⟨0.6,0.4⟩ | ⟨0.9,0.3⟩ | ⟨0.6,0.7⟩ | ⟨0.1,0.7⟩ |
| $E_2$ | $A_1$ | ⟨0.3,0.9⟩ | ⟨0.7,0.6⟩ | ⟨0.5,0.8⟩ | ⟨0.3,0.6⟩ | ⟨0.3,0.6⟩ |
| | $A_2$ | ⟨0.7,0.4⟩ | ⟨0.9,0.2⟩ | ⟨0.8,0.1⟩ | ⟨0.3,0.5⟩ | ⟨0.3,0.5⟩ |
| | $A_3$ | ⟨0.3,0.6⟩ | ⟨0.7,0.7⟩ | ⟨0.7,0.6⟩ | ⟨0.4,0.4⟩ | ⟨0.4,0.3⟩ |
| | $A_4$ | ⟨0.4,0.8⟩ | ⟨0.7,0.5⟩ | ⟨0.6,0.2⟩ | ⟨0.4,0.7⟩ | ⟨0.4,0.7⟩ |
| | $A_5$ | ⟨0.2,0.7⟩ | ⟨0.8,0.2⟩ | ⟨0.8,0.4⟩ | ⟨0.6,0.6⟩ | ⟨0.6,0.6⟩ |
| $E_3$ | $A_1$ | ⟨0.6,0.8⟩ | ⟨0.7,0.6⟩ | ⟨0.5,0.8⟩ | ⟨0.5,0.5⟩ | ⟨0.1,0.6⟩ |
| | $A_2$ | ⟨0.6,0.5⟩ | ⟨0.9,0.2⟩ | ⟨0.8,0.1⟩ | ⟨0.3,0.5⟩ | ⟨0.3,0.4⟩ |
| | $A_3$ | ⟨0.4,0.7⟩ | ⟨0.7,0.5⟩ | ⟨0.6,0.1⟩ | ⟨0.2,0.9⟩ | ⟨0.6,0.5⟩ |
| | $A_4$ | ⟨0.2,0.9⟩ | ⟨0.5,0.6⟩ | ⟨0.6,0.2⟩ | ⟨0.1,0.6⟩ | ⟨0.4,0.7⟩ |
| | $A_5$ | ⟨0.1,0.6⟩ | ⟨0.8,0.2⟩ | ⟨0.9,0.2⟩ | ⟨0.6,0.5⟩ | ⟨0.4,0.6⟩ |

Table 2: Pythagorean fuzzy group normalized decision matrix

| Alternatives | $C_1$ | $C_2$ | $C_3$ | $C_4$ | $C_5$ |
|---|---|---|---|---|---|
| $A_1$ | ⟨0.45,0.83⟩ | ⟨0.73,0.6⟩ | ⟨0.54,0.77⟩ | ⟨0.38,0.64⟩ | ⟨0.34,0.6⟩ |
| $A_2$ | ⟨0.67,0.47⟩ | ⟨0.9,0.2⟩ | ⟨0.8,0.3⟩ | ⟨0.3,0.54⟩ | ⟨0.42,0.47⟩ |
| $A_3$ | ⟨0.34,0.58⟩ | ⟨0.6,0.64⟩ | ⟨0.67,0.42⟩ | ⟨0.43,0.61⟩ | ⟨0.48,0.44⟩ |
| $A_4$ | ⟨0.43,0.78⟩ | ⟨0.64,0.54⟩ | ⟨0.63,0.2⟩ | ⟨0.33,0.64⟩ | ⟨0.37,0.7⟩ |
| $A_5$ | ⟨0.32,0.67⟩ | ⟨0.74,0.28⟩ | ⟨0.87,0.31⟩ | ⟨0.6,0.6⟩ | ⟨0.42,0.64⟩ |

Table 3: Arithmetic Average of Pythagorean fuzzy decision matrix

| Alternatives | $C_1$ | $C_2$ | $C_3$ | $C_4$ | $C_5$ |
|---|---|---|---|---|---|
| $A_1$ | 0.16 | 0.07 | 0.09 | 0.19 | 0.24 |
| $A_2$ | 0.08 | 0.0 | 0.2 | 0.06 | 0.18 |
| $A_3$ | 0.18 | 0.3 | 0.33 | 0.37 | 0.16 |
| $A_4$ | 0.26 | 0.15 | 0.07 | 0.23 | 0.07 |
| $A_5$ | 0.23 | 0.18 | 0.12 | 0.1 | 0.32 |

Table 4: Maximum radius lengths based on decision matrices

| Alternatives | $C_1$ | $C_2$ | $C_3$ | $C_4$ | $C_5$ |
|---|---|---|---|---|---|
| $A_1$ | ⟨0.45,0.83; 0.16⟩ | ⟨0.73,0.6; 0.07⟩ | ⟨0.54,0.77; 0.09⟩ | ⟨0.38,0.64; 0.19⟩ | ⟨0.34,0.6; 0.24⟩ |
| $A_2$ | ⟨0.67,0.47; 0.08⟩ | ⟨0.9,0.2; 0.0⟩ | ⟨0.8,0.3; 0.2⟩ | ⟨0.3,0.54; 0.06⟩ | ⟨0.42,0.47; 0.18⟩ |
| $A_3$ | ⟨0.34,0.58; 0.18⟩ | ⟨0.6,0.64; 0.3⟩ | ⟨0.67,0.42; 0.33⟩ | ⟨0.43,0.61; 0.37⟩ | ⟨0.48,0.44; 0.16⟩ |
| $A_4$ | ⟨0.43,0.78; 0.26⟩ | ⟨0.64,0.54; 0.15⟩ | ⟨0.63,0.2; 0.07⟩ | ⟨0.33,0.64; 0.23⟩ | ⟨0.37,0.7; 0.07⟩ |
| $A_5$ | ⟨0.32,0.67; 0.23⟩ | ⟨0.74,0.28; 0.18⟩ | ⟨0.87,0.31; 0.12⟩ | ⟨0.6,0.6; 0.1⟩ | ⟨0.42,0.64; 0.32⟩ |

Table 5: Circular Pythagorean fuzzy decision matrix

**Step 4:** The decision matrix expressed with C-PFVs for each alternatives are aggregated by utilizing aggregation operators $CPWA_q^A$, $CPWA_p^A$, $CPWG_q^A$ and $CPWG_p^A$ defined via $g(t) = -\log t^2$, $g(t) = -\log(1-t^2)$, $q(t) = -\log t^2$ and $p(t) = -\log(1-t^2)$ in Remark 4 and Remark 5. Aggregated circular Pythagorean fuzzy decision matrix for each aggregation operators is shown in Table 6.

| Alternatives | $CPWA_q^A$ | $CPWA_p^A$ | $CPWG_q^A$ | $CPWG_p^A$ |
|---|---|---|---|---|
| $A_1$ | ⟨0.59,0.66; 0.11⟩ | ⟨0.59,0.66; 0.13⟩ | ⟨0.52,0.69; 0.11⟩ | ⟨0.52,0.69; 0.13⟩ |
| $A_2$ | ⟨0.78,0.32; 0.0⟩ | ⟨0.78,0.32; 0.11⟩ | ⟨0.65,0.38; 0.0⟩ | ⟨0.65,0.38; 0.11⟩ |
| $A_3$ | ⟨0.53,0.55; 0.25⟩ | ⟨0.53,0.55; 0.27⟩ | ⟨0.5,0.57; 0.25⟩ | ⟨0.5,0.57; 0.27⟩ |
| $A_4$ | ⟨0.54,0.56; 0.14⟩ | ⟨0.54,0.56; 0.17⟩ | ⟨0.49,0.63; 0.14⟩ | ⟨0.49,0.63; 0.17⟩ |
| $A_5$ | ⟨0.66,0.43; 0.18⟩ | ⟨0.66,0.43; 0.22⟩ | ⟨0.56,0.52; 0.18⟩ | ⟨0.56,0.52; 0.22⟩ |

Table 6: Aggregated Circular Pythagorean fuzzy decision matrix

**Step 5:** The cosine similarity measure $CSM$ defined in Definiton 17 is used to measure how each aggregated C-PFV and positive ideal alternative are related or closed to each other. The results of similarity measure between positive ideal alternative and alternatives is shown in Table 7.

| | $CSM(A_1, A^+)$ | $CSM(A_2, A^+)$ | $CSM(A_3, A^+)$ | $CSM(A_4, A^+)$ | $CSM(A_5, A^+)$ |
|---|---|---|---|---|---|
| $CPWA_q^A$ | 0.325 | 0.493 | 0.465 | 0.411 | **0.555** |
| $CPWA_p^A$ | 0.377 | 0.548 | 0.475 | 0.425 | **0.571** |
| $CPWG_q^A$ | 0.301 | **0.473** | 0.429 | 0.328 | 0.468 |
| $CPWG_p^A$ | 0.311 | **0.528** | 0.439 | 0.343 | 0.488 |

Table 7: The results of similarity measure between positive ideal alternative and alternatives

**Step 6:** With respect to the aggregation operators $CPWA_q^A$ and $CPWA_p^A$ we get the ranking $A_1 \prec A_4 \prec A_3 \prec A_2 \prec A_5$ and with respect to the aggregation operators $CPWG_q^A$ and $CPWG_p^A$ we get the ranking $A_1 \prec A_4 \prec A_3 \prec A_5 \prec A_2$. The steps of the proposed method are visualized in Figure 6.

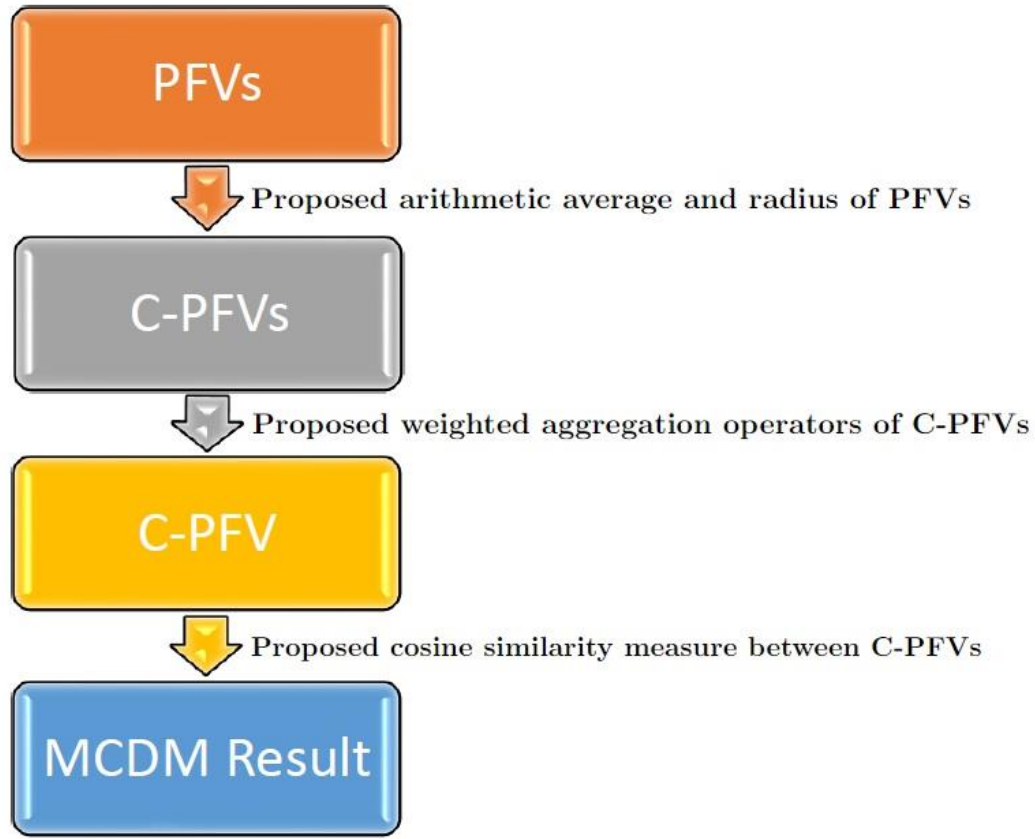

Figure 6: Application of the proposed method to MCDM

### 5.4 Comparative analysis

The best alternative remains same with the literature when the aggregation operators $CPWG_q^A$ and $CPWG_p^A$ are used. On the other hand the orders of best and second best alternative interchange when the aggregation operators $CPWA_q^A$ and $CPWA_p^A$ are used. The worst alternative is totally consistent with the literature. The comparison of the other methods proposed to solve this MCDM problem and the method we propose is shown in Table 8 and illustrated in Figure 7.

| Methods | Ranking order | Best Alternative |
|---|---|---|
| Zhang [39] | $A_1 < A_3 < A_4 < A_5 < A_2$ | $A_2$ |
| Biswas and Sarkar [7] | $A_1 < A_4 < A_3 < A_5 < A_2$ | $A_2$ |
| Biswas and Sarkar [8] | $A_1 < A_3 < A_4 < A_5 < A_2$ | $A_2$ |
| Proposed Method ($CPWA_q^A$) | $A_1 < A_4 < A_3 < A_2 < A_5$ | $A_5$ |
| Proposed Method ($CPWA_p^A$) | $A_1 < A_4 < A_3 < A_2 < A_5$ | $A_5$ |
| Proposed Method ($CPWG_q^A$) | $A_1 < A_4 < A_3 < A_5 < A_2$ | $A_2$ |
| Proposed Method ($CPWG_p^A$) | $A_1 < A_4 < A_3 < A_5 < A_2$ | $A_2$ |

Table 8: The comparison of the other methods and proposed method

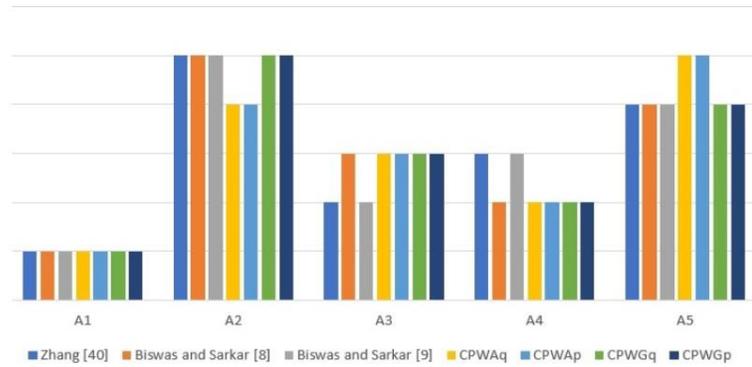

Figure 7: The column chart comparison of the other methods and proposed method

### 5.5 Time complexity of the proposed MCDM method

In this sub-section we investigate the time complexity of the MCDM method given in Sub-section 5.2. We assume that $m$ experts assign PFVs to create the decision matrix as in the problem solved in Sub-section 5.3. Essentially the time complexity that depends on the number of times of multiplication, exponential, summation as in [9] and [13] is evaluated. Consider a MCDM problem with $n$ alternatives, k criteria and $m$ experts. In Step 2 we need $k + 2knm$ operations, in Step 3 we need $nk(10m + 6)$ operations, in Step 4 we need $n(8k + 2)$ operations if we utilize the aggregation operator $CPWA_q^A$, we need $n(10k + 4)$ operations if we utilize the aggregation operator $CPWA_p^A$, we need $n(8k + 2)$ operations if we utilize the aggregation operator $CPWG_q^A$ and we need $n(10k + 4)$ operations if we utilize the aggregation operator $CPWG_p^A$. In Step 5, we need $23n$ operations. Therefore the time complexity is

$$Comp(k, n; m) = k + 2kn(6m + 7) + 25n$$

for the aggregation operators $CPWA_q^A$ and $CPWG_q^A$,

$$Comp(k,n;m) = k + 4kn(3m+4) + 27n$$

for the aggregation operators $CPWA_p^A$ and $CPWG_p^A$. Obviously the bi-variate functions $g_q, g_p: [2, \infty) \to \mathbb{R}$ defined by $g_m^q(x,y) = x + 2xy(6m+7) + 25y$ and $g_m^p(x,y) = x + 4xy(3m+4) + 27y$ assumes global minimum at point $(2,2)$ for any fixed positive integer $m$. Figure 8 visualizes the change of the time complexity with respect to the change in the numbers of the criteria, the alternatives and the experts for $CPWA_q^A$, $CPWA_p^A$, $CPWG_q^A$ and $CPWG_p^A$.

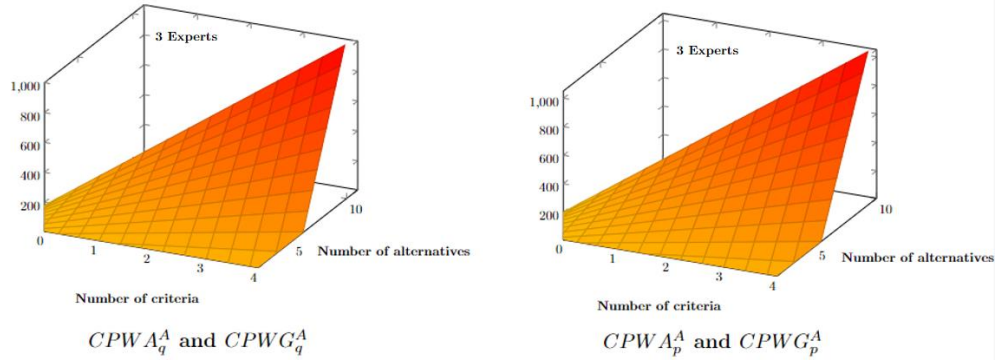

Figure 8: Time complexity of the proposed MCDM method

## 6 Conclusion

The main goal of this paper is to introduce the concept of C-PFS represented by a circle whose radius is $r$ and whose center consists is a pair with the condition that sum of the square of the components is less than one. In such a fuzzy set the membership degree and the non-membership degree are represented by a circle. Thus a C-PFS is a generalization of both C-IFSs and PFSs. C-PFSs allow decision makers or experts to evaluate objects in a larger and more flexible region compared to both C-IFSs and PFSs. Therefore, the change of membership degree and non-membership degree can be handled to express uncertainty with the help of C-PFSs. In this way, more sensitive decisions can be made. In this paper, a method is developed to transform PFVs to a C-PFS. Also some fundamental set theoretic operations for C-PFSs are given and some algebraic operations for C-PFVs via continuous Archimedean $t$-norms and $t$-conorms are introduced. Then with the help of these algebraic operations some weighted aggregation operators for C-PFVs are presented. Inspired by a cosine similarity measure defined between PFVs, we give a cosine similarity measure based on radius to determine the degree of similarity between C-PFVs. Finally, by utilizing the concepts mentioned above we propose a MCDM method in circular Pythagorean fuzzy environment and we apply the proposed method to a MCDM problem from the literature about selecting the best photovoltaic cell (also known as solar cell). We compare the results of the proposed method with the existing results and calculate the time complexity of the MCDM method. In the future studies, different kind of aggregation operators and similarity measures can be investigated. Also while transforming PFVs to a C-PFS other aggregation tool as fuzzy integrals or aggregation operators can be used. Moreover, the proposed method can be used to solve MCDM problems such as classification, pattern recognition, data mining, clustering and medical diagnosis.